\documentclass{article}

\usepackage{iclr2027_conference,times}

\usepackage[utf8]{inputenc}
\usepackage[T1]{fontenc}
\usepackage{hyperref}
\usepackage{url}
\usepackage{booktabs}
\usepackage{multirow}
\usepackage{amsfonts}
\usepackage{amsmath}
\usepackage{amssymb}
\usepackage{svg}
\usepackage{nicefrac}
\usepackage{microtype}
\usepackage{xcolor}
\usepackage{graphicx}
\usepackage{algorithm}
\usepackage{algorithmic}
\usepackage{subcaption}
\usepackage{wrapfig}
\usepackage{bm}

\newcommand{\R}{\mathbb{R}}
\newcommand{\dout}{{d_{\mathrm{out}}}}
\newcommand{\din}{{d_{\mathrm{in}}}}
\newcommand{\dhidden}{{d_{\mathrm{h}}}}
\newcommand{\lap}{\nabla^2}
\newcommand{\vibenet}{\texttt{beignet}}

\newcommand{\vct}[1]{{\bm{#1}}}
\newcommand{\mat}[1]{\mathbf{#1}}

\title{Fourier Feature Pyramids \\ for Physics-Informed Neural Networks}

\author{%
\begin{tabular}[t]{@{}l@{}}
\bf Brandon Zhao$^{1*}$ \quad \bf Yixuan Wang$^{1}$ \quad \bf Jonathan T. Barron$^{3}$ \quad \bf Katherine L. Bouman$^{1,2}$ \\[0.12cm]
\bf Dor Verbin$^{3}$ \quad \bf Pratul P. Srinivasan$^{3}$ \\[0.25cm]
\normalfont $^1$Department of Computing and Mathematical Sciences, California Institute of Technology\\
\normalfont $^2$Cahill Center for Astronomy and Astrophysics, California Institute of Technology\\
\normalfont $^3$Google DeepMind\\
{\tt\small $^*$byzhao@caltech.edu}
\end{tabular}%
}

\iclrfinalcopy 
\begin{document}

\maketitle

\begin{abstract}
We present an improved neural field architecture for solving partial differential equations (PDEs). Current physics-informed neural networks (PINNs) provide a flexible framework for solving PDEs, but they struggle to achieve highly accurate solutions and require computation that scales poorly with parameter count. Our model, which we call \vibenet{} (Bandlimited Embedding with Interpolated Grid Network), replaces the random Fourier feature embedding used by existing PINN models with a trainable multi-resolution Fourier feature pyramid. To query \vibenet{} at a continuous coordinate, we use Fourier interpolation at each level of the pyramid to return features at the input coordinate, and then decode this vector with a fully-connected neural network trunk. Our model provides multiple benefits: 1) Spatial derivatives can be computed efficiently by using the chain rule to compose derivatives of the neural network computed with automatic differentiation with derivatives of the feature grid computed spectrally by the Fast Fourier transform (FFT). 2) \vibenet{} can achieve higher accuracy in a compute-efficient manner by scaling the parameter count of this Fourier feature pyramid, instead of the less-efficient strategy of scaling the neural network architecture. 3) \vibenet{} can directly control the representation bandlimit, resulting in more stable optimization for difficult PDEs. We demonstrate that \vibenet{} finds significantly more accurate solutions on PDE benchmarks using fewer parameters than state-of-the-art PINN methods. We further evaluate \vibenet{} on the self-similar inviscid Burgers blowup problem and show that it can minimize residuals to near machine precision using Adam, an accuracy regime previously attained only by using computationally expensive higher-order optimizers.
\end{abstract}

\section{Introduction}
\label{sec:intro}

Physics-informed neural networks (PINNs)~\citep{raissi2019physics} are a flexible approach for approximating partial differential equation (PDE) solutions by parameterizing them as coordinate-based neural networks (multilayer perceptrons, or MLPs, that take in coordinates and output values of the field). PINNs are optimized to minimize the PDE residual over batches of coordinates sampled from the solution domain.
As noted in prior work~\citep{basri2020frequency,rahaman2019spectral,xu2019frequency,xu2019training}, neural networks suffer from spectral bias as standard gradient descent optimization dynamics result in them learning low-frequency components much faster than high-frequency components.
Fourier feature embeddings~\citep{tancik2020fourier} mitigate this problem by passing input coordinates through a set of sinusoidal mappings before applying the MLP. In PINNs, the frequencies for these sinusoids are typically sampled from Gaussian distributions, with a single scalar that controls the bandwidth of the induced kernel~\citep{wang2021eigenvector}.

Existing approaches for scaling PINNs to more challenging problems rely on increasing either the Fourier feature bandwidth or the size of the underlying MLP, but both strategies introduce significant optimization and computational challenges. Because Gaussian Fourier features are not strictly bandlimited, increasing the Gaussian bandwidth can introduce higher-frequency components into the induced kernel that can make optimization brittle. Alternatively, increasing the size of the MLP substantially increases the computational cost of evaluating PDE derivatives through automatic differentiation. The computational burden of scaling MLP-based PINNs is exacerbated by the need for second-order optimization methods to achieve high numerical accuracy~\citep{wang2026high,wang2025discovery}, and therefore memory consumption becomes prohibitive for even moderately large architectures.

We introduce a PINN architecture, which we call \vibenet{}, that addresses these issues by replacing Gaussian random
Fourier features with a learnable multi-resolution Fourier feature pyramid. In the spatial domain, this representation can be thought of as a set of periodic feature grids at coarse-to-fine spatial resolutions. To evaluate the field at any continuous coordinate, we simply use Fourier interpolation to compute a single feature for each level of the pyramid, stack these into a feature vector for the coordinate, and pass this vector to a small MLP decoder. This representation allows us to bandlimit the representation by limiting the maximum resolution in the pyramid and also benefit from the coarse-to-fine optimization with a multiscale representation. Furthermore, the equispaced grid structure also enables efficient computation: feature values and derivatives can be evaluated on structured grids using the Fast Fourier transform (FFT) and then propagated through an MLP via the chain rule.

\vibenet{}'s explicit control over spatial frequency content is particularly useful for PINNs because different PDEs have different spectral requirements. Nonlinear terms in a PDE can generate high harmonics or sharp spatial structure, favoring broader frequency support. At the same time, differential operators amplify high frequency errors, so indiscriminately increasing the bandwidth can result in unstable optimization. Our Fourier feature pyramid exposes these choices directly: the finest grid resolution determines the maximum spatial frequency represented in the feature embedding, while the number of levels and channels per level controls the representational capacity within that band.

We evaluate \vibenet{} in two regimes. First, using four time-dependent PDEs from the JAX-PI benchmark~\citep{wang2023expert}, we compare against the JAX-PI and PirateNet PINN architectures using a common evaluation protocol. These benchmarks span both scalar 1D+time PDEs and two-component 2D+time reaction--diffusion systems with periodic spatial domains. Across all four tasks, \vibenet{} achieves substantially lower relative error. We also investigate how different PDEs favor different spectral parameterizations---some benefit from broader frequency support (e.g., Allen--Cahn) while others perform best with a more compact bandwidth and greater concentration of representational capacity in low- and mid-frequency bands (e.g., KdV)---and demonstrate how \vibenet{}'s design space is well-suited for navigating this tradeoff. 
Second, we study how \vibenet{} can support extreme-precision optimization on a self-similar inviscid Burgers blowup problem, an important stepping stone toward understanding singularity formation in PDEs. This broader class of questions is central to mathematical fluid dynamics: the unforced Navier--Stokes existence and smoothness problem, for example, remains among the most fundamental open problems in mathematical analysis~\citep{fefferman2006existence}. In this setting, coordinate-based MLPs trained with Adam plateau at much higher
residuals, while \vibenet{} drives the fixed-grid residuals to near machine precision with the same first-order optimizer. These results suggest
that the feature representation is a key ingredient for scaling PINNs to high-accuracy regimes.

\section{Related Work}
\paragraph{Spectral-bias remedies in neural representations.}
Standard MLPs tend to fit smooth low-frequency structure before fine high-frequency
variation. This behavior has motivated neural field and PINN architectures that
make high-frequency components easier to represent or optimize. Fourier feature
embeddings~\citep{tancik2020fourier}, reviewed in
Section~\ref{sec:fourier_features}, address this by mapping input coordinates with sinusoids of varying frequency. SIREN~\citep{sitzmann2020siren} uses sinusoidal activation functions to represent fine detail and obtain accurate derivatives,
BACON~\citep{lindell2022bacon} constructs band-limited coordinate networks with
a multiscale representation, and WIRE~\citep{saragadam2023wire} uses wavelet
activations to combine frequency localization with spatial localization. Another line of work modifies the network parameterization through learnable univariate functions:
Kolmogorov--Arnold Networks (KANs)~\citep{liu2024kan,liu2024kan2} have been
studied as a way to improve expressivity and mitigate spectral
bias~\citep{wang2024expressiveness}, with extensions to
PINNs~\citep{shukla2024comprehensive,toscano2025pinns} and neural
operators~\citep{lee2025kano}. Our work shares the goal of improving
high-frequency representation in PINNs, but instead of modifying the MLP decoder,  we focus on improving the input feature representation.

\paragraph{Fourier-basis PINN architectures.}
Several works have incorporated Fourier bases into PINNs beyond the random
Fourier feature embeddings discussed in Section~\ref{sec:fourier_features}.
\citet{wang2024piratenets} discussed equispaced spectral features as a coordinate embedding in their ablation studies, reporting that this achieved higher training accuracy than Random Fourier Features (RFF) but exhibited overfitting. They consequently adopted RFF in their method's architecture. Our work revisits this design choice: by organizing equispaced
features into a learnable multi-resolution pyramid and using stochastic grid shifts as regularization, we make deterministic equispaced
Fourier features effective in the PINN setting while retaining direct control
over spatial bandwidth and frequency-band redundancy. In a complementary architectural direction, Fourier PINNs~\citep{cooley2024fourier}
combine equispaced Fourier bases additively with an MLP output and train via alternating least squares
with $L^2$-regularized basis pruning. Our architecture instead prepends the
equispaced basis as an input representation, preserving compatibility with the
dominant architecture paradigm (RFF followed by a neural network decoder) and enabling direct substitution into existing architectures. 
Spectral-Informed Neural Networks~\citep{yu2025spectral} operate entirely in the spectral domain, taking
frequencies as inputs and outputting Fourier coefficients for PDEs with
periodic boundary conditions, a different regime from coordinate-based PINNs. 

\paragraph{Fourier neural operators.}
Fourier Neural Operators (FNOs)~\citep{kovachki2023neural,li2020fourier}
also use spectral representations for PDEs, but typically target a different
learning setting: they learn an amortized solution operator across many PDE
instances, while PINNs optimize a representation for a particular instance from
residual, boundary, and initial-condition losses. Physics-Informed Neural
Operators (PINO)~\citep{li2021physics,maust2022fourier} narrow this gap by
adding PDE residual losses to operator learning. Architecturally, FNOs use
repeated grid-to-grid spectral convolution layers, whereas \vibenet{} uses a
multi-resolution latent feature grid that is interpolated to collocation points
and decoded by a small MLP.

\section{Background}

\subsection{Physics-informed neural networks for solving PDEs}

Consider a general partial differential equation (PDE) parameterized by a differential operator $\mathcal{F}$ acting on an unknown function $u:\R^\din\to\R^\dout$:
\begin{equation} \label{eq:pde}
    \mathcal{F}(\vct{x}, u, \nabla_\vct{x} u, \nabla_\vct{x}^2u, ...) = 0,
\end{equation}
with some boundary conditions.

Physics-informed neural networks (PINNs) provide a framework to numerically approximate the solution $u$ using a neural network. The neural network is trained to approximate a PDE solution by minimizing the square of the residual on the left hand side of Equation~\ref{eq:pde}, averaged over batches of collocation points in the domain.

\subsection{Spectral Bias and the Neural Tangent Kernel}
\label{sec:spectral_bias_ntk}

Coordinate-based neural networks trained with gradient descent learn
low-frequency components of a target function before high-frequency
ones~\citep{basri2020frequency,rahaman2019spectral}. This \emph{spectral bias}
is a property of optimization rather than expressivity, and it is especially
relevant for PINNs, whose solutions may contain sharp interfaces, localized
waves, or other fine-scale structure.

The neural tangent kernel (NTK) formalizes this phenomenon \citep{jacot2018ntk}. For a network
$f_\vct{\theta}$ with training inputs $\{\vct{x}_i\}_{i=1}^N$, the empirical NTK
at initialization is $\mat{K}_{ij} = \langle \nabla_\vct{\theta}
f_{\vct{\theta}_0}(\vct{x}_i), \nabla_\vct{\theta}
f_{\vct{\theta}_0}(\vct{x}_j) \rangle$. In the infinite-width or
linearized-training regime, $\mat{K}$ remains approximately constant, and each
component of the training error along an eigenvector of $\mat{K}$ decays
exponentially under gradient flow at a rate proportional to the corresponding
eigenvalue. When the kernel is stationary and the inputs lie on a uniform periodic grid, the eigenvectors of $\mat{K}$ are Fourier modes and the eigenvalue $\lambda(\mathbf{k})$ governs the convergence rate of
the frequency-$\mathbf{k}$ error component. Spectral bias then corresponds to
$\lambda(\mathbf{k})$, and consequently the convergence rate in frequency $\mathbf{k}$, decaying rapidly with $|\mathbf{k}|$ in coordinate MLPs. Shaping the kernel spectrum
$\lambda(\mathbf{k})$ therefore controls how quickly each frequency component
of the solution is learned.

\subsection{Fourier Feature Embeddings}
\label{sec:fourier_features}

Fourier feature embeddings address spectral bias by modifying the input representation so that the NTK allocates more eigenvalue mass to higher spatial frequencies, thereby accelerating their optimization.
A Fourier feature embedding maps coordinates to a collection of sinusoidal
features before applying a neural network~\citep{tancik2020fourier}. Given a
frequency matrix $\mat{B}$ with $M$ rows, the embedding is
\begin{equation}
    \gamma(\vct{x})
    =
    \begin{bmatrix}
        \cos(\mat{B} \vct{x})\\
        \sin(\mat{B} \vct{x})
    \end{bmatrix}\,,
    \label{eq:fourier_embedding}
\end{equation}
with the trigonometric functions applied element-wise. A Fourier feature network
then takes the form $f_\vct{\theta}(\vct{x})=\mathrm{MLP}_\vct{\theta}(\gamma(\vct{x}))$. In the case of a linear Fourier feature model $f(\vct{x})=\frac{1}{\sqrt{M}}\mat{W}\gamma(\vct{x})$,
with fixed matrix $\mat{B}$ and trainable output weights $\mat{W}$, the NTK is exactly the feature kernel
\begin{equation}
    \mat{K}(\vct{x},\vct{x}')
    =
    \frac{1}{M}
    \sum_{j=1}^{M}
    \cos\!\left(\vct{b}_j^\top(\vct{x}-\vct{x}')\right)\,,
    \label{eq:linear_fourier_feature_kernel}
\end{equation}
where $\vct{b}_j$ is the $j$th row of $\mat{B}$. Thus, the frequencies in $\mat{B}$ determine the spectrum of the kernel. For a nonlinear MLP,
the NTK composes the MLP kernel with this feature kernel; in the infinite-width limit it remains stationary, and its spectrum spreads around each embedded frequency with a smooth falloff~\citep{tancik2020fourier}. As a result, one can approximately tune the convergence behavior of a Fourier feature network by adjusting the frequencies in $\mat{B}$.

Most prior work constructs $\mat{B}$ using random Fourier features (RFFs), typically sampling entries of $\mat{B}$ independently from a Gaussian distribution
$\mathcal{N}(0,\sigma^2)$. This is motivated by the classical kernel setting, where Gaussian RFFs give a
Monte Carlo approximation to a stationary RBF
kernel~\citep{rahimi2007random, rahimi2008weighted}, whose bandwidth is set by $\sigma$~\citep{tancik2020fourier}. This simple bandwidth
control has made RFFs a standard tool for PINNs, where they substantially
improve on coordinate networks without a feature
embedding~\citep{tancik2020fourier, wang2021eigenvector}. 
Gaussian draws concentrate near low frequencies, so higher-frequency components of the solution are covered sparsely and converge slowly. Such components matter little at moderate accuracy but can be limiting in high-precision regimes.
\vibenet{} instead places its frequencies on a multi-resolution grid, which covers equispaced frequencies up to a chosen cutoff and lets us set how densely each band is covered.


\section{Method}
\label{sec:method}

Our goal is to design a parametric family of neural function approximators which can be used to numerically solve a wide range of PDEs. Our representation, \vibenet{}, is based on Fourier features but we use a fixed multi-resolution grid of frequencies instead of random sampling. Our design enables tight control of the spectral content of the resulting function while enabling efficient computation of the output and its derivatives.

Section~\ref{sec:feature_pyramid} describes our multi-resolution Fourier pyramid, the fundamental component in our method. Section~\ref{sec:temporal_lerp} describes the extension of our method to time-dependent PDEs: each pyramid level stores a small number of learned temporal feature slices, which are linearly interpolated and decoded by a single shared MLP. Section~\ref{sec:spectral_feature_derivatives} details how our representation enables efficient computation of derivatives, and Section~\ref{sec:stochastic_shift} describes how we use the same properties to reduce aliasing.

\begin{figure}[t]
  \centering
  \includegraphics[width=\linewidth]{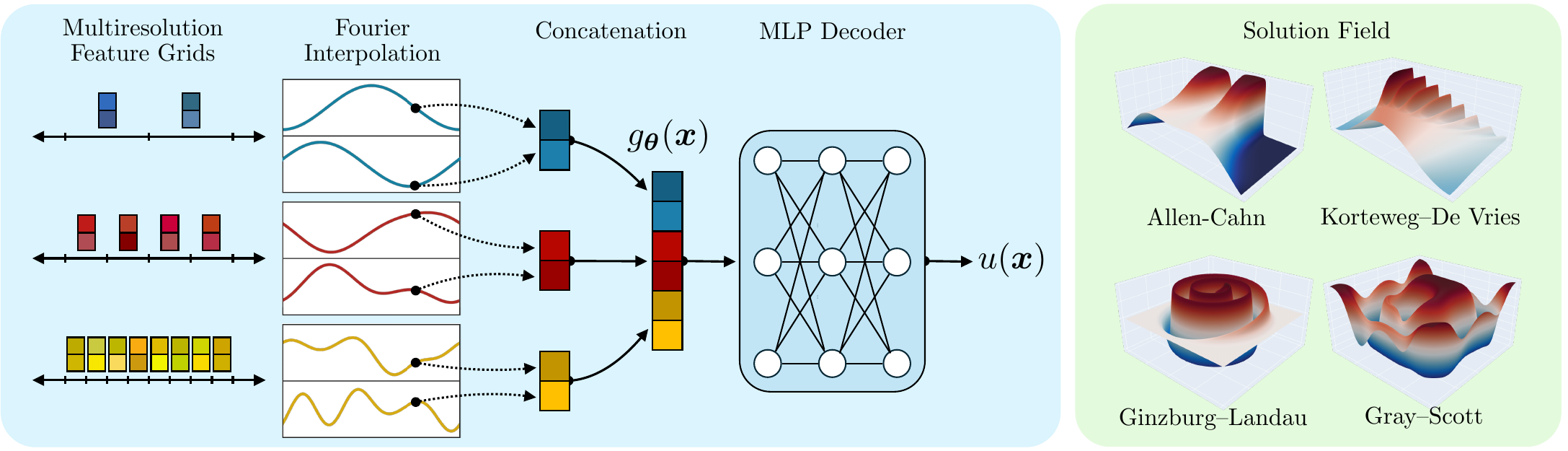}
  \caption{Architecture overview for \vibenet{}. A learnable multi-resolution
  Fourier feature pyramid stores feature grids at coarse-to-fine spatial
  resolutions, inducing a corresponding hierarchy of bandlimited feature
  fields. For each query point, every level is evaluated by Fourier
  interpolation to produce a per-level feature vector. These feature vectors are
  concatenated and decoded by a compact MLP to produce the solution value at a particular input coordinate; for
  time-dependent problems, the normalized time coordinate is also passed to the
  decoder. Our proposed architecture \vibenet{} is supervised on the initial conditions and PDE residual, and can be evaluated at arbitrary points inside the domain. \vibenet{} achieves SOTA performance on four time-dependent PDEs from the JAX-PI suite \citep{wang2023expert}, detailed in Sec. \ref{sec:experiments}.}
  \label{fig:architecture}
\end{figure}

\subsection{Multi-resolution Fourier feature pyramid}
\label{sec:feature_pyramid}

Our architecture relies on a learnable multi-resolution pyramid of periodic spatial feature grids. Each level of the pyramid stores a feature field at a prescribed grid resolution. Evaluating the model at a point consists of two steps. First, we use Fourier interpolation to query the input coordinate at each of the model's levels. We concatenate the resulting features, then pass them through a decoder MLP.

Formally, we represent our function $u : \R^\din \to \mathbb{R}^{\dout}$ as the composition of $L$ Fourier interpolation operations $g_{\vct{\theta}^{(\ell)}} : \R^\din\to\R^\dhidden$ with parameters $\vct{\theta}^{(\ell)}$, one at each level $\ell = 1, ..., L$, followed by an MLP $f_\vct{\phi}: \R^{L\cdot \dhidden} \to\R^\dout$ with parameters $\vct{\phi}$:
\begin{equation}
    u(\vct{x}) = f_\vct{\phi}\left(g_{\vct{\theta}^{(1)}}(\vct{x}),g_{\vct{\theta}^{(2)}}(\vct{x}), ..., g_{\vct{\theta}^{(L)}}(\vct{x})  \right)\,.
\end{equation}

Let $\widehat{\vct{\theta}}^{(\ell)}$ denote the normalized discrete Fourier transform of
the learned spatial feature grid $\vct{\theta}^{(\ell)}$. The Fourier
interpolation operation at the $\ell$th level can then be written as
\begin{equation} \label{eq:ferp}
    g_{\vct{\theta}^{(\ell)}}(\vct{x})
    =
    \mathrm{Re}\Bigl(
    \sum_{\mathbf{k}\in \mathcal{K}^{(\ell)}}
    \widehat{\vct{\theta}}^{(\ell)}[\mathbf{k}]
    \exp\!\left(2\pi i\, \mathbf{k}\cdot \vct{x}\right)
    \Bigr)\,,
\end{equation}
where $\widehat{\vct{\theta}}^{(\ell)}[\mathbf{k}]$ is the
$\dhidden$-dimensional Fourier coefficient vector for mode $\mathbf{k}$.
For a grid
with resolution $N^{(\ell)}$ (assumed to be even) the
signed FFT mode set is:
\begin{equation}
    \mathcal{K}^{(\ell)} = 
    \left\{-\frac{N^{(\ell)}}{2}, -\frac{N^{(\ell)}}{2}+1,...,\frac{N^{(\ell)}}{2}-1
    \right\}^\din\,.
\end{equation}

The resolution $N^{(\ell)}$ determines the spectral support available at that  level, while the channel dimension $\dhidden$ controls how much feature capacity is allocated within that support.
By changing the number of levels $L$, their resolutions $N^{(1)}, ..., N^{(L)}$, and their channel counts $\dhidden$, the model can allocate more or less capacity to different bands. Unlike prior work which uses random Fourier features, this is done without relying on a random frequency draw. In \vibenet{}'s multi-resolution parameterization, the grid resolutions determine which feature-level frequencies are available, while the level structure and channel counts determine how much trainable feature capacity is placed at each scale. In practice, we choose the resolutions to scale dyadically, e.g.,
$N^{(\ell)} = 2^{\ell-1} N^{(1)}$, in the spirit of classical
image-pyramid constructions such as the Laplacian pyramid
\citep{burt1983laplacian}; however, the parameterization itself allows
arbitrary choices of grid resolutions $N^{(1)}, \ldots, N^{(L)}$.

\subsection{Temporal Pyramid Blending}
\label{sec:temporal_lerp}

For time-dependent PDEs, the solution can vary sharply in space while evolving
non-periodically in time. Naïvely treating time as an additional coordinate and applying the same Fourier construction for it would impose an artificial temporal periodicity and couple the spatial FFT resolution to the number of time samples.

Our solution is to divide the time axis $t \in [0, 1]$ into $T$ bins (in all of our time-dependent PDE experiments we use $T=7$, i.e., $8$ temporal slices), where for $i=0, ..., T-1$, the $i$th bin spans $[\nicefrac{i}{T}, \nicefrac{i+1}{T})$. We store $T+1$ pyramids, $\vct{\theta}_0, ..., \vct{\theta}_{T}$, each consisting of $L$ levels, and in order to evaluate the function at a specific time value $t$, we linearly interpolate between the features:
\begin{equation}
  \vct{\theta}(t) = \vct{\theta}_i + \left(tT - i\right)(\vct{\theta}_{i+1} - \vct{\theta}_i),
\end{equation}
where $i = \min(\lfloor tT \rfloor, T-1)$ is the index of the bin containing $t$, and the interpolation is done identically over all elements in all levels of the pyramids. The interpolated pyramid is then used for Fourier interpolation as in Equation~\ref{eq:ferp}, and fed into the shared MLP to predict the output value. In practice, we also input the time coordinate $t$ directly to the MLP.

Our representation allows the spatial representation to remain bandlimited, while treating time as a non-periodic coordinate. The learned temporal slices provide local time-varying features, while the raw time coordinate gives the MLP a direct continuous variable for smooth trends and for chaining time derivatives through the decoder. The number of time bins $T$ controls the number of learned temporal anchors in the pyramid.
\vspace{-0.1in}
\subsection{Spectral Feature Derivatives}
\label{sec:spectral_feature_derivatives}

The Fourier-interpolated feature grids also provide an efficient way to compute
the spatial derivatives needed for computing the PDE residual (see Equation~\ref{eq:pde}). We do this by differentiating the feature fields in Fourier space and then propagating through the decoder by the chain rule.

For a pyramid level $\ell$, let
$\vct{\theta}^{(\ell)}$ be the temporally interpolated spatial
feature grid at time $t$ (see Section~\ref{sec:temporal_lerp} for details on this interpolation). 
Let $\widehat{\vct{\theta}}^{(\ell)}$ denote the normalized discrete Fourier transform of $\vct{\theta}^{(\ell)}$.
The spatial derivatives of the feature field
$g_{\vct{\theta}^{(\ell)}}$ can be computed in Fourier space by multiplying
these coefficients by the wavenumbers. This can be seen by taking the derivative of Equation~\ref{eq:ferp}:
\begin{equation}
    \nabla_\vct{x}g_{\vct{\theta}^{(\ell)}}(\vct{x}) =
    \mathrm{Re}\Bigl(
    \sum_{\mathbf{k}\in \mathcal{K}^{(\ell)}}
    \widehat{\vct{\theta}}^{(\ell)}[\mathbf{k}] (2\pi i \mathbf{k})
    \exp\!\left(2\pi i\, \mathbf{k}\cdot \vct{x}\right)
    \Bigr),
\end{equation}
and this generalizes to higher-order derivatives by multiplying by the corresponding higher-order wavenumber factors.

After multiplication by the appropriate wavenumber factors, we use the inverse Fast Fourier Transform (FFT) to obtain feature values and spatial derivative features on the structured collocation grid. When needed, spectra are zero-padded to match the residual evaluation grid before the inverse FFT. Raw step-time comparisons for this FFT-structured residual path are given in Appendix~\ref{app:ac_timing_efficiency}.

This process allows us to compute all spatial derivatives of the Fourier interpolation $g$ defined in Equation~\ref{eq:ferp}, but our goal is to compute the derivatives of the entire function $u$, i.e., the Fourier interpolation composed with the MLP. To do that, we use Faà di Bruno's formula, which generalizes the chain rule to higher order derivatives. For example, the first derivative of $u$ is simply:
\begin{equation}
    \nabla_{\vct{x}} u(\vct{x}) = \sum_{\ell=1}^{L} \frac{\partial f}{\partial g_{\vct{\theta}^{(\ell)}}} \nabla_{\vct{x}} g_{\vct{\theta}^{(\ell)}}(\vct{x})\,.
\end{equation}

In
implementation, these terms are evaluated with nested forward-mode automatic
differentiation, using the spectrally computed feature derivatives as tangent
directions. This avoids forming dense derivative tensors while still computing
the pointwise derivatives required by the PDE residual.

The same construction applies componentwise in multiple spatial dimensions,
with mixed derivatives obtained by multiplying Fourier coefficients by the
corresponding products of wavenumbers. Thus, the PDE residual is evaluated by
combining FFT-based feature derivatives with chain-rule derivatives through the
decoder. This preserves the bandlimited structure of each pyramid level while
allowing the nonlinear decoder to mix features across scales.

\subsection{Stochastic grid shift}
\label{sec:stochastic_shift}

Fixed collocation grids create a potential issue with aliasing: the network can learn high-frequency functions that satisfy the PDE residual at grid points while remaining inaccurate between them. This is particularly relevant for Fourier feature grids, which make high-frequency feature fields directly available to the MLP.
To reduce this effect, we randomly offset the resolution $M$ structured residual grid during training by a random shift $\boldsymbol{\delta} \sim \mathrm{Uniform}([0,1/M]^\din)$ (assuming for simplicity that the inputs are in $[0, 1]^\din$). The shift is implemented efficiently as a phase ramp in Fourier space, multiplying the component of the wavenumber $\mathbf{k}$ by $\exp\!\left(2\pi i \mathbf{k}\cdot\,\boldsymbol{\delta}\right)$.
This preserves the structured FFT evaluation while training on off-grid spatial locations; we ablate this choice in Appendix~\ref{app:stochastic_shift_results}.

\section{Experiments}
\label{sec:experiments}
\subsection{Time-dependent PDE benchmarks}
\label{sec:time_dependent}

We evaluate \vibenet{} on a benchmark containing four time-dependent PDEs from the
JAX-PI suite~\citep{wang2023expert}: Allen--Cahn, Korteweg--de Vries (KdV), complex
Ginzburg--Landau, and Gray--Scott. In all benchmark experiments, we replace the random Fourier feature embedding in
JAX-PI with the \vibenet{} Fourier feature pyramid while keeping the ModifiedMLP backbone architecture (with width reduced to 128 for the two 2D systems); exact configurations are provided in the supplement. These problems span a wide range of behaviors and different dimensionalities of the domain and range of the function $u$ being approximated, including mapping from 1-dimensional spatial coordinates and time to a scalar field, mapping from 2-dimensional spatial coordinates and time to a vector field, reaction--diffusion equations, and dispersive wave equations. While all four equations are periodic, which makes them natural for FFT-based feature derivatives, periodicity is not essential to our approach; we show results on non-periodic Dirichlet problems with irregular geometry in Appendix \ref{app:irregular_geometry}.

The benchmarks test different aspects of the representation. Allen--Cahn
requires resolving sharp transition layers generated by nonlinear
reaction--diffusion dynamics. KdV contains a third-order dispersive derivative,
which makes broad high-frequency spatial representations difficult to train.
Complex Ginzburg--Landau and Gray--Scott extend the evaluation to coupled
two-field 2D systems with fine-scale pattern formation. We generally match the training strategy 
of the JAX-PI benchmark suite. Benchmark details and \vibenet{} configurations are given in Appendix~\ref{app:vibenet_configurations}, and solution visualizations in Appendix~\ref{app:solution_visualizations}.

\begin{table}[t]
  \caption{
  Time-dependent PDE comparison. We report relative $L_2$ error on the
  primary field and, for two-component systems, on the secondary field, as
  mean $\pm$ standard deviation over three random seeds (42, 43, 44). For
  Ginzburg--Landau and Gray--Scott, the
  table reports the per-window parameter count with the number of windows.
  JAX-PI and PirateNet results are reproduced from the JAX-PI codebase under
  the same evaluation protocol.
  }
  \label{tab:time_dependent}
  \centering
  \small
  \resizebox{\linewidth}{!}{%
  \begin{tabular}{llccc}
    \toprule
    Problem & Method & $L_2$ error & $L_2$ error (sec.) & Params \\
    \midrule
    \multirow{3}{*}{Allen--Cahn}
      & JAX-PI \citep{wang2023expert}& $(6.98 \pm 5.39)\times10^{-5}$ & --- & 396,930 \\
      & PirateNet \citep{wang2024piratenets}         & $(1.21 \pm 0.58)\times10^{-5}$ & --- & 1,916,041 \\
      & \vibenet{} (ours) & $\mathbf{(1.88 \pm 0.22)\times10^{-6}}$ & --- & 411{,}938 \\
    \midrule
    \multirow{3}{*}{KdV}
      & JAX-PI \citep{wang2023expert}            & $(4.06 \pm 1.05)\times10^{-4}$ & --- & 396,930 \\
      & PirateNet \citep{wang2024piratenets}        & $(2.59 \pm 0.15)\times10^{-4}$ & --- & 1,916,041 \\
      & \vibenet{} (ours) & $\mathbf{(3.71 \pm 0.63)\times10^{-5}}$ & --- & 412{,}674 \\
    \midrule
    \multirow{3}{*}{Ginzburg--Landau}
      & JAX-PI \citep{wang2023expert}            & $(2.10 \pm 0.44)\times10^{-2}$ & $(2.04 \pm 0.96)\times10^{-2}$ & $331{,}396 \times 5$ \\
      & PirateNet \citep{wang2024piratenets}         & $(1.15 \pm 0.43)\times10^{-2}$ & $(9.41 \pm 4.97)\times10^{-3}$ & $1{,}916{,}553 \times 5$ \\
      & \vibenet{} (ours) & $\mathbf{(1.08 \pm 0.16)\times10^{-3}}$ & $\mathbf{(1.31 \pm 0.55)\times10^{-3}}$ & $855{,}936 \times 5$ \\
    \midrule
    \multirow{3}{*}{Gray--Scott}
      & JAX-PI \citep{wang2023expert}            & $(7.28 \pm 6.98)\times10^{1}$ & $(8.08 \pm 12.95)\times10^{2}$ & $331{,}396 \times 10$ \\
      & PirateNet \citep{wang2024piratenets}        & $(3.10 \pm 0.14)\times10^{-3}$ & $(8.00 \pm 0.41)\times10^{-3}$ & $1{,}916{,}553 \times 10$ \\
      & \vibenet{} (ours) & $\mathbf{(1.84 \pm 1.52)\times10^{-4}}$ & $\mathbf{(4.80 \pm 3.95)\times10^{-4}}$ & $855{,}936 \times 10$ \\
    \bottomrule
  \end{tabular}}
\end{table}

As shown in Table \ref{tab:time_dependent}, \vibenet{} finds significantly higher accuracy solutions than both JAX-PI and PirateNet across all four benchmarks.
The improvement is largest on Ginzburg--Landau and Gray--Scott, where the solution dynamics place
strong demands on the spatial representation. On Ginzburg--Landau, \vibenet{}
reduces error by roughly an order of magnitude relative to PirateNet.
On Gray--Scott, \vibenet{} reduces the error by more than an order of magnitude
on both fields while using less than half the per-window parameter count of
PirateNet. These results support the central premise of the method: allocating
spectral capacity in the feature representation can be a more parameter-efficient
scaling axis than increasing only the MLP backbone.

\subsection{Self-similar blowup in 1D Burgers}
\label{sec:burgers_blowup}

A promising application of high-accuracy PINNs is the computation of
self-similar singularity profiles.  In singularity formation problems, the
quantity of interest may become unbounded in finite time, making direct
time-dependent simulation increasingly ill-conditioned near blowup.  A common
strategy is instead to introduce self-similar variables, which transform the
search for a finite-time singularity into the search for a stationary profile \citep{hou2014self}.
Accurate numerical profiles can then serve as candidates for subsequent
stability analysis or computer-assisted proof.  This direction has recently
received renewed attention in fluid singularity problems, beginning with neural
network computations of asymptotic self-similar profiles for the 3D
axisymmetric Euler equations~\citep{wang2023asymptotic}, and continuing in 1D
model equations, Boussinesq-type systems, and related fluid settings
\citep{wang2026high,wang2025discovery}. Developing methods capable of accurately resolving such singular structures is an important step toward the broader challenge of understanding singularity formation in fluid dynamics, including the questions underlying the unforced Navier--Stokes Millennium Prize Problem \citep{fefferman2006existence}.

This setting is a demanding test of optimization accuracy. Recent
high-precision PINN approaches~\citep{wang2026high,wang2025discovery} can
identify self-similar profiles, but rely on expensive second-order or
Gauss--Newton-type refinement to reach residual levels suitable for downstream
analysis. This may limit scalability in higher dimensions, where richer
representations are needed and curvature-based methods can scale quadratically
or worse with parameter count. Here, we ask whether the Fourier feature pyramid
can reach comparable precision using first-order optimization alone.

Following \citet{wang2026high}, we use the 1D inviscid Burgers equation as a model problem. Although simple, it retains the key ingredients needed for our investigation: a self-similar blowup ansatz, an unbounded profile domain, and a residual minimization problem whose attainable precision is highly sensitive to the choice of neural representation. We consider the 1D Burgers equation:

\begin{equation}  \label{eq:burgers}
    u_t + u u_x = 0,
\end{equation}
and seek self-similar solutions of the form:
\begin{equation}
    u(x,t) = \tau^\lambda U(y),
    \qquad
    y=x\tau^{-(1+\lambda)},
    \qquad
    \tau=T-t,
    \label{eq:burgers_ansatz}
\end{equation}
where $\lambda$ denotes the similarity exponent, $T$ is the blowup time, and $U(y)$ is a stationary profile that describes the rescaled solution shape near the singularity. Rather than learning the full time-dependent solution $u(x,t)$, the PINN is trained to represent the profile function $U(y)$. Substituting this ansatz into the PDE in Equation~\ref{eq:burgers} gives the stationary profile equation:
\begin{equation}
    -\lambda U
    +
    \bigl((1+\lambda)y+U\bigr)U_y
    =
    0 .
    \label{eq:burgers_profile}
\end{equation}
In the experiments below we fix
$\lambda=0.5$ and minimize the squared profile residual together with the
derivative residual used by \citet{wang2026high}. Implementation details for the profile representation, domain mapping, and
residual terms are provided in the supplement.

\begin{table}[t]
  \caption{
    Self-similar 1D Burgers profile with fixed $\lambda=0.5$.  This controlled
    singularity-profile problem tests whether the Fourier feature pyramid can move
    first-order PINN training into the high-precision regime.  Using only Adam,
    \vibenet{} reaches near-machine precision residual accuracy, while the Adam-trained coordinate MLP
    baseline remains many orders of magnitude less accurate.  This suggests that
    representation design can reduce reliance on expensive second-order refinement
    for high-precision profile computations.  Memory scaling refers to optimizer
    state as model size grows; Adam is linear in parameters, while full BFGS
    stores a dense inverse-Hessian approximation and is quadratic.
    \vspace{1mm}}
  \label{tab:burgers}
  \centering
  \small
  \begin{tabular}{lcccc}
    \toprule
    Method & Optimizer & Memory & PDE MSE & $\log_{10}\max |F_U|$ \\
    \midrule
    MLP \citep{wang2026high} & Adam
      & $\mathcal{O}(P)$
      & $6.21 \times 10^{-7}$
      & $-2.41$ \\
    MLP \citep{wang2026high} & BFGS
      & $\mathcal{O}(P^2)$
      & $7.15 \times 10^{-20}$
      & $-9.10$ \\
    \vibenet{} pyramid (12$\times$4) & Adam
      & $\mathcal{O}(P)$
      & $6.29 \times 10^{-19}$
      & $-8.46$ \\
    \bottomrule
  \end{tabular}
\end{table}

Table~\ref{tab:burgers} compares the residual accuracy obtained by a coordinate
MLP and by the Fourier feature pyramid on the fixed diagnostic grid.  With Adam
alone, the coordinate MLP stalls at a PDE MSE of $6.21\times10^{-7}$, whereas
the 12-level \vibenet{} pyramid reaches $6.29\times10^{-19}$ after the same
20K-step schedule.  The maximum pointwise profile residual shows a comparable
gap, improving from $\log_{10}\max |F_U|=-2.41$ for the coordinate MLP to
$-8.46$ for \vibenet{}. The learned profiles and pointwise residuals are shown in
Figure~\ref{fig:solutions_burgers}.

This experiment complements the time-dependent PDE benchmarks above.  There,
\vibenet{} improves accuracy and parameter efficiency on standard PINN tasks;
here, in a low-dimensional singularity-profile problem, the same representation
substantially changes the residual floor reachable by first-order optimizers like Adam. This makes
Fourier feature pyramids a promising direction for higher-dimensional
self-similar profile computations, where second-order refinement with second-order optimizers like BFGS can become substantially more expensive.

\subsection{PDE-dependent spectral allocation}
\label{sec:pyramid_tuning}

\begin{figure}[t]
  \centering
  \includegraphics[width=\linewidth]{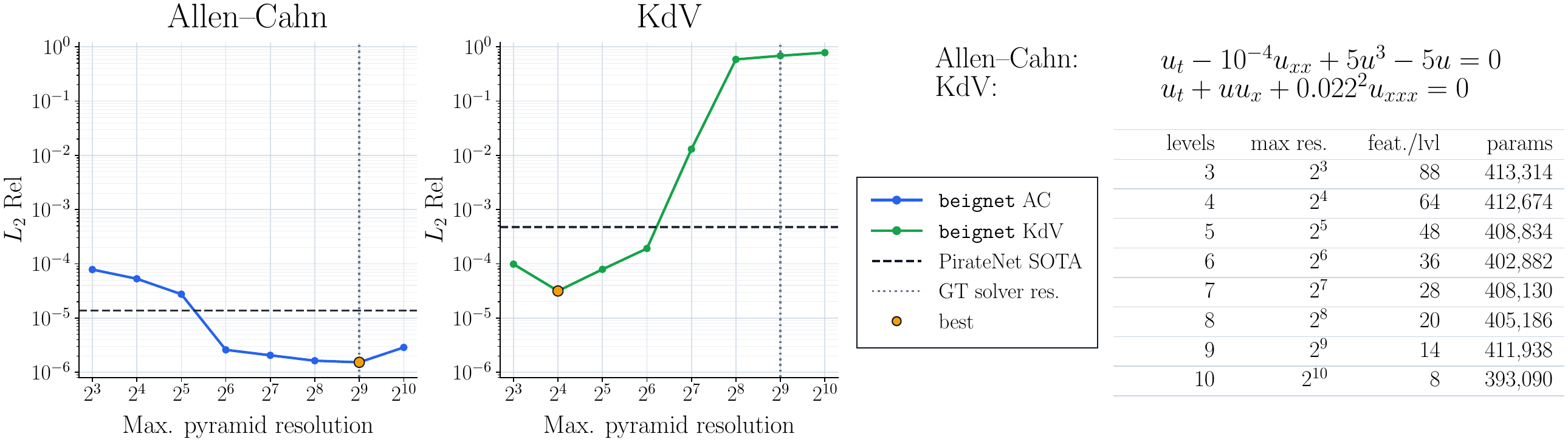}
  \caption{
  Feature-pyramid size sweep for Allen--Cahn and KdV. Each point is a
  parameter-matched \vibenet{} model in which the number of pyramid levels sets
  the maximum represented grid resolution and the number of features per level
  is adjusted to keep the total parameter count approximately fixed. Dashed
  horizontal lines show PirateNet's error (single run) relative to the numerical reference solution, and the vertical dotted line indicates the spatial resolution of the numerical solver used to generate that reference solution. Allen--Cahn benefits from increasing pyramid bandwidth up to
  the resolution of the numerical solver, whereas KdV is most accurate with a lower-bandwidth pyramid.
  }
  \label{fig:pyramid_size_sweep}
\end{figure}

The central design choice in \vibenet{} is not only the overall support of the Fourier features, but how capacity is distributed across resolutions. Unlike
random Fourier features, whose frequency coverage is randomly drawn from a single-parameter distribution, \vibenet{} allocates features across all the chosen grid resolutions. This exposes two practical choices: the maximum resolution of
the finest grid, which sets the representation's frequency support, and the allocation of
feature channels across levels, which sets the decoder budget within that
frequency support. We choose these quantities with a small parameter-matched sweep. In
Figure~\ref{fig:pyramid_size_sweep}, all models have roughly $4\times 10^5$
parameters; moving right increases the maximum grid resolution while reducing
the number of features per level.

Interestingly, the two one-dimensional benchmarks favor different allocations. Allen--Cahn
continues to improve as the pyramid adds finer levels, reaching its best error
near the $2^9$ reference-grid resolution. This is expected because its solutions
can contain sharp interfaces, so fine features help represent real structure in
the solution; at the same time, the diffusive term damps small high-frequency
errors. KdV behaves differently. It contains a third spatial derivative, which weights every Fourier
mode by a factor proportional to its frequency cubed. This makes the weight of high-frequency components in the residual extremely high, which can make optimization harder, a standard
source of stiffness in KdV time-stepping methods \citep{klein2008fourth}. Thus,
adding very fine feature levels can hurt even when the parameter count is held
fixed. The best KdV model is the shallow $4$-level, $64$-feature pyramid, which
keeps more capacity in low and middle frequencies rather than allocating it to
very fine scales. This sweep motivates treating the pyramid bandwidth as a
PDE-dependent design parameter rather than a monotone scaling knob.
\vspace{-0.1in}

\section{Conclusion}
\label{sec:conclusion}

We have introduced \vibenet{}, a PINN architecture that replaces fixed Gaussian
random Fourier features with a deterministic multi-resolution Fourier feature
pyramid. The central idea is to treat the feature embedding as a spectral
allocation: the pyramid controls both the spatial bandwidth of the
representation and the redundancy assigned to each frequency band. This provides
a simple way to adapt Fourier feature capacity to the PDE structure, retaining broad coverage when high frequencies are useful while avoiding
excessive capacity in bands that cause training instability.

On four time-dependent JAX-PI benchmarks, \vibenet{} improves over the JAX-PI and
PirateNet baselines. On a self-similar PDE blowup problem, our representation enables Adam to
drive the PDE residual near machine precision, multiple orders of magnitude lower than what is achievable with Adam-trained versions of the standard coordinate-based MLPs used for this problem setting.
These results suggest that carefully designing the feature representation itself is a key ingredient for pushing PINNs into new high-accuracy regimes, opening the door to neural PDE solvers capable of resolving fine-scale structure, extreme multiscale dynamics, and challenging phenomena such as singularities.

\subsection*{AI use statement}
Generative AI tools were used to assist with implementing the method, running
experiments, and drafting paper text. Generative AI was not used for any other task requiring
disclosure under the ICLR 2027 policy. The authors manually verified all
AI-generated code and results and substantially revised all AI-drafted text.
The authors take full responsibility for the content of this paper.

\subsection*{Reproducibility statement}
Our code is available at
\url{https://github.com/beignet-paper/beignet-release/}. It includes the
\vibenet{} implementation and scripts for all experiments in this paper.
Per-problem \vibenet{} configurations are given in
Appendix~\ref{app:vibenet_configurations}, the JAX-PI benchmark setup and
evaluation protocol in Appendix~\ref{app:jaxpi_benchmarks}, and the
self-similar Burgers blowup setup in Appendix~\ref{app:burgers_blowup_details}.

\ificlrfinal
\subsubsection*{Acknowledgments}
The authors would like to thank Matthieu Darcy for helpful discussions. This work was supported by the Amazon AI4Science Partnership Discovery Grant,  Carver Mead New Adventures Fund, Ginsburg Prize, and Sloan Fellowship. 
\fi

\bibliographystyle{iclr2027_conference}
\bibliography{references}

\clearpage
\appendix

\section{Solution Visualizations}
\label{app:solution_visualizations}

\begin{figure}[h]
  \centering
  \includegraphics[width=\linewidth]{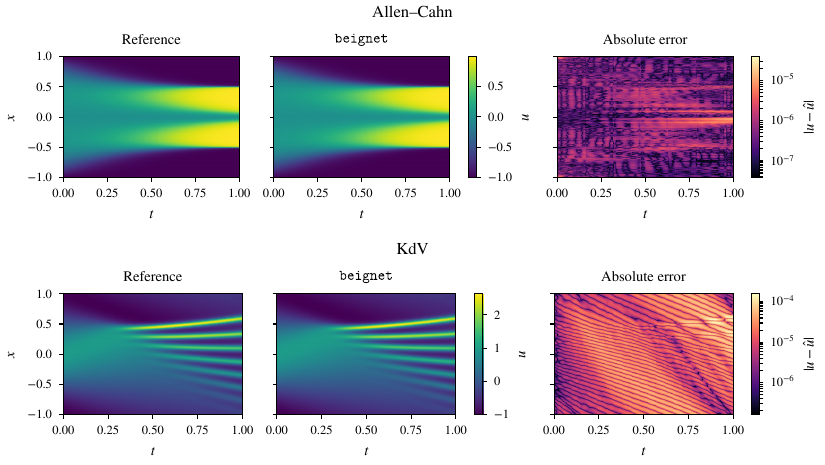}
  \caption{
    Allen--Cahn (top) and KdV (bottom) solutions over $(t,x)$: reference
    solution, \vibenet{} prediction, and absolute error (log scale). The shown runs have relative $L_2$ errors of $1.62\times10^{-6}$ (Allen--Cahn)
    and $3.40\times10^{-5}$ (KdV).
  }
  \label{fig:solutions_1d}
\end{figure}

\begin{figure}[h]
  \centering
  \includegraphics[width=\linewidth]{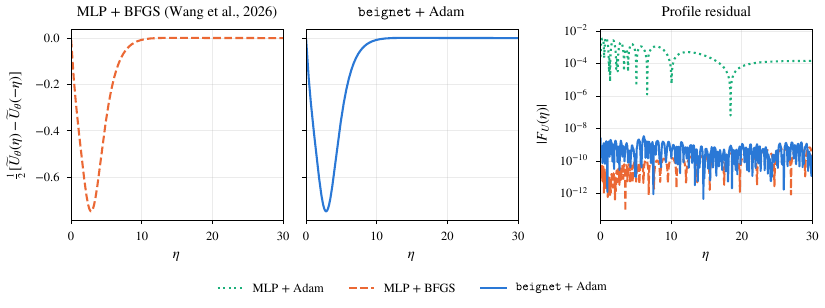}
  \caption{
    Self-similar Burgers profile with $\lambda=0.5$ on the diagnostic grid of
    Appendix~\ref{app:burgers_blowup_details}. Left and middle: the learned odd
    component of Eq.~\eqref{eq:app_burgers_repr} for the coordinate MLP trained
    with BFGS~\citep{wang2026high} and for \vibenet{} trained with Adam. Right:
    pointwise profile residual $|F_U|$ from Eq.~\eqref{eq:app_burgers_residual}
    for all three runs of Table~\ref{tab:burgers}, including the coordinate MLP
    trained with Adam.
  }
  \label{fig:solutions_burgers}
\end{figure}

\begin{figure}[p]
  \centering
  \includegraphics[width=\linewidth]{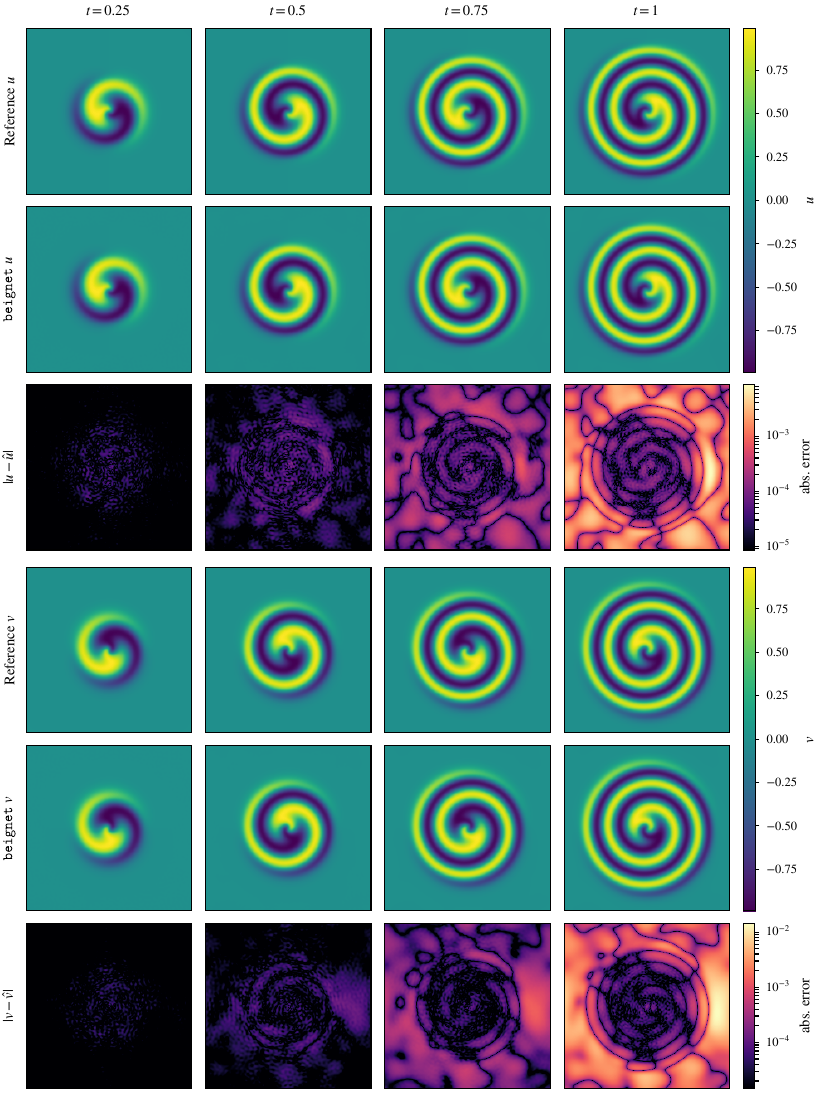}
  \caption{
    Complex Ginzburg--Landau snapshots on $[-1,1]^2$ at
    $t\in\{0.25,0.5,0.75,1\}$. Rows show the reference solution, the \vibenet{}
    prediction, and the absolute error (log scale) for $u$ (top three rows) and
    $v$ (bottom three rows). The displayed run has full-trajectory relative
    $L_2$ errors of $1.04\times10^{-3}$ ($u$) and $1.94\times10^{-3}$ ($v$).
  }
  \label{fig:solutions_ginzburg_landau}
\end{figure}

\begin{figure}[p]
  \centering
  \includegraphics[width=\linewidth]{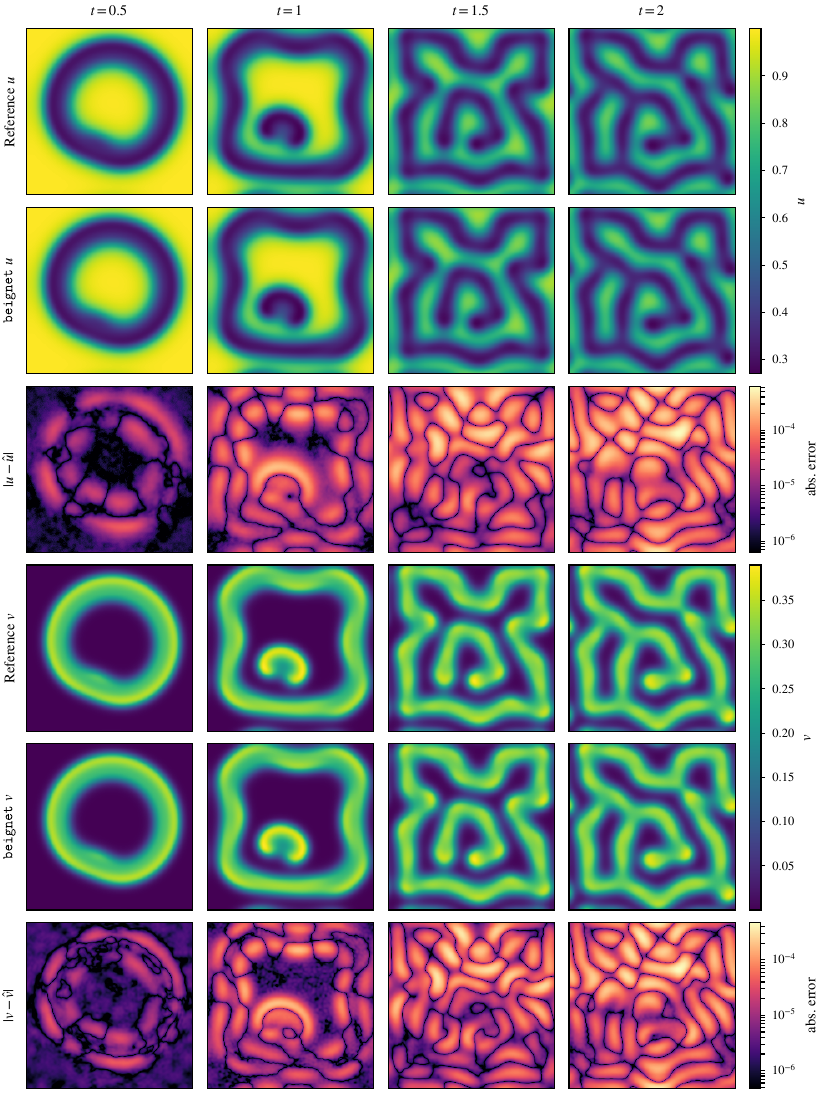}
  \caption{
    Gray--Scott snapshots on $[-1,1]^2$ at $t\in\{0.5,1,1.5,2\}$. Rows show the
    reference solution, the \vibenet{} prediction, and the absolute error (log
    scale) for $u$ (top three rows) and $v$ (bottom three rows). The displayed run has full-trajectory relative $L_2$ errors of
    $9.95\times10^{-5}$ ($u$) and $2.60\times10^{-4}$ ($v$).
  }
  \label{fig:solutions_grey_scott}
\end{figure}

We visualize the \vibenet{} solutions for Allen--Cahn and KdV in
Figure~\ref{fig:solutions_1d}, Ginzburg--Landau in
Figure~\ref{fig:solutions_ginzburg_landau}, and Gray--Scott in
Figure~\ref{fig:solutions_grey_scott}. Figure~\ref{fig:solutions_burgers} shows
the self-similar Burgers profile and its residual, and
Figure~\ref{fig:solutions_pinnacle} shows the irregular-geometry Poisson problems
of Appendix~\ref{app:irregular_geometry}. Absolute errors are plotted on a
logarithmic color scale spanning three decades below the largest error shown.

\begin{figure}[h]
  \centering
  \includegraphics[width=\linewidth]{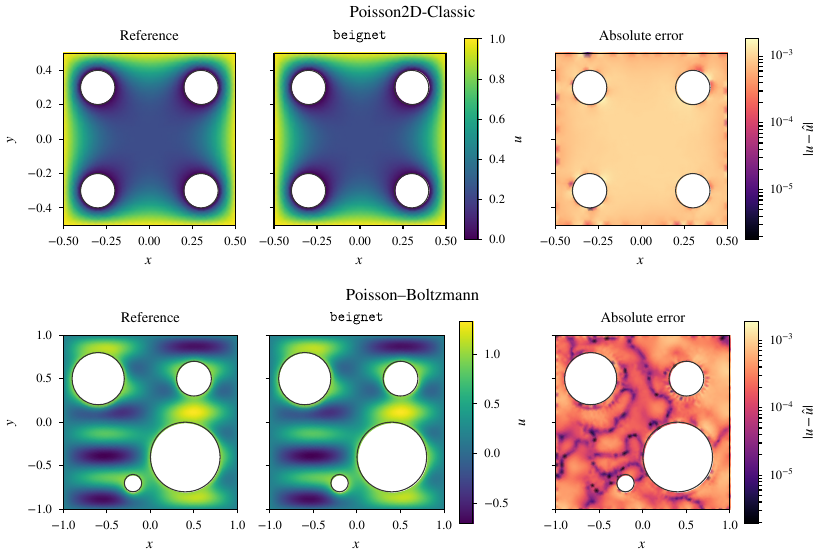}
  \caption{
    Poisson2D-Classic (top) and Poisson--Boltzmann (bottom) from the PINNacle benchmark \citep{hao2024pinnacle} on their irregular
    domains: COMSOL reference solution, \vibenet{} prediction, and absolute error
    (log scale), shown on the COMSOL mesh nodes with linear interpolation
    between nodes. The displayed runs have relative $L_2$ errors of
    $1.81\times10^{-3}$ (Poisson2D-Classic) and $6.25\times10^{-4}$
    (Poisson--Boltzmann).
  }
  \label{fig:solutions_pinnacle}
\end{figure}

\clearpage
\section{Non-periodic domains with irregular geometry}
\label{app:irregular_geometry}

We test \vibenet{} on two Dirichlet problems with irregular geometry from
PINNacle~\citep{hao2024pinnacle}: Poisson2D-Classic ($-\lap u = 0$) and
Poisson--Boltzmann ($-\lap u + 64u = f$), each posed on a square with four
circular holes and Dirichlet data on all eight boundary components. We embed the
domain in a periodic box twice as wide in each direction, enforce the PDE
residual only at stochastically shifted lattice points inside the domain, and
impose the boundary conditions with penalty terms. We compare against an MLP and
an MLP with Gaussian random Fourier features ($\sigma = 1$), both taking raw
coordinates and parameter-matched to \vibenet{} (7 levels up to $128^2$). All models share
the same training settings and are evaluated by relative $L_2$ error on the
COMSOL reference nodes provided with PINNacle. \vibenet{} attains the lowest
error on both problems (Table~\ref{tab:irregular_geometry}), $2.4\times$ and
$5.7\times$ lower than the best baseline, respectively; the solutions are
visualized in Figure~\ref{fig:solutions_pinnacle}.

\begin{table}[h]
\centering
\caption{
Dirichlet problems with irregular geometry from PINNacle: relative $L_2$ error,
mean $\pm$ standard deviation over three random seeds (42, 43, 44).
}
\label{tab:irregular_geometry}
\small
\begin{tabular}{llcc}
\toprule
Problem & Method & Relative $L_2$ error & Params \\
\midrule
\multirow{3}{*}{Poisson2D-Classic}
  & MLP & $(4.04 \pm 1.15)\times10^{-3}$ & 85,261 \\
  & RFF + MLP & $(3.70 \pm 2.62)\times10^{-3}$ & 85,667 \\
  & \vibenet{} (ours) & $\mathbf{(1.56 \pm 0.26)\times10^{-3}}$ & 85,689 \\
\midrule
\multirow{3}{*}{Poisson--Boltzmann}
  & MLP & $(6.68 \pm 2.58)\times10^{-3}$ & 131,041 \\
  & RFF + MLP & $(4.12 \pm 0.65)\times10^{-3}$ & 131,023 \\
  & \vibenet{} (ours) & $\mathbf{(7.22 \pm 3.99)\times10^{-4}}$ & 130,777 \\
\bottomrule
\end{tabular}
\end{table}

\section{Image Fitting}
\label{app:image_fitting}

Although \vibenet{} is designed for PDE residual training, the underlying
Fourier feature pyramid is a general coordinate-regression representation. To
isolate this representation from the PDE-specific residual and derivative
machinery, we also evaluate it on direct RGB image fitting, following the image
regression setting used by \citet{tancik2020fourier} to study Fourier feature
embeddings.

We fit a starfish image by training a coordinate network
$(x,y)\mapsto (r,g,b)$ with a full-grid RGB mean-squared-error objective on a $256 \times 256$ starfish image. All models are trained for 2000 Adam steps with learning rate
$10^{-3}$. We compare a vanilla coordinate MLP, a Gaussian random Fourier
feature model with $\sigma=10$, and our learned Fourier feature pyramid. The vanilla MLP and RFF baselines use a larger $256$-wide,
$4$-layer decoder, while \vibenet{} uses a smaller $64$-wide, $3$-layer
decoder after its learned multi-resolution feature pyramid. Despite this
smaller decoder, \vibenet{} achieves near-perfect reconstruction and uses fewer
trainable parameters than the number of RGB values in the image itself.

\begin{figure}[t]
  \centering
  \includegraphics[width=\linewidth]{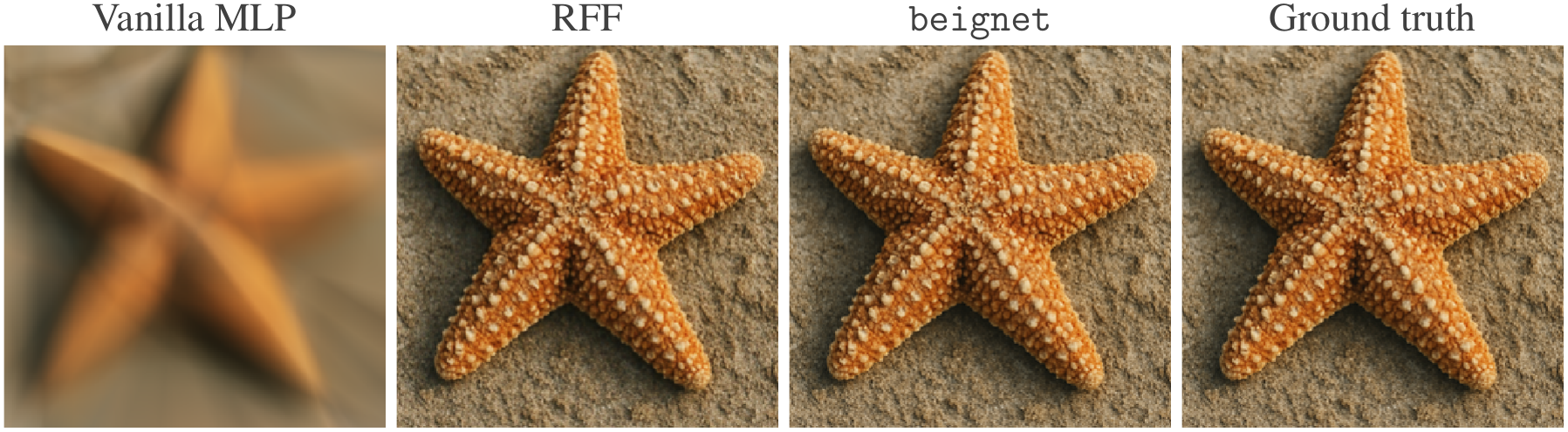}
  \caption{
    Direct $256\times256$ RGB image fitting.
    The \vibenet{} representation fits the image nearly exactly under the same
    2000-step optimization budget, while using a smaller decoder and fewer
    trainable parameters than the number of stored RGB values in the target image.
    }
  \label{fig:starfish_image_fitting}
\end{figure}

\begin{table}[h]
\centering
\caption{
Starfish image-fitting parameter count and PSNR at resolution $256\times256$.
The ground-truth row counts the number of stored RGB
image values, $256\times256\times3$. 
}
\label{tab:starfish_image_fitting}
\small
\begin{tabular}{@{}lcc@{}}
\toprule
Method & Params & PSNR (dB) \\
\midrule
Vanilla MLP & 133,123 & 19.61 \\
RFF, $\sigma=10$ & 264,195 & 36.73 \\
\vibenet{} & 180,203 & 58.39 \\
Ground truth image & 196,608 & --- \\
\bottomrule
\end{tabular}
\end{table}

\section{Ablation Studies}
We include three ablations to isolate the contributions of stochastic grid
shifts, the coarse-to-fine pyramid structure, and temporal pyramid blending.
Unless otherwise stated, all ablations are trained with the same physics-informed
objective used in the main experiments, including the PDE residual loss and the
corresponding initial and boundary condition terms. The reported PDE residual is
a training loss quantity evaluated on collocation points, whereas the relative
$L_2$ error is a test loss measured against a reference numerical solution on
the evaluation grid. This distinction is important for interpreting the
ablations: a small residual on the training collocation grid does not necessarily
imply small solution error between those points.

\subsection{Stochastic Grid-Shift prevents fixed-grid overfitting}
\label{app:stochastic_shift_results}

Table~\ref{tab:shift_ablation} reports experimental results with and without stochastic shifting. The effect depends on the pyramid shape: Allen--Cahn uses a wide pyramid reaching spatial resolution $512$, so the
unshifted grid can overfit fixed collocation locations while generalizing poorly between them (low PDE residual but high relative $L_2$ error). KdV uses a much shallower pyramid, with capacity concentrated in
low and mid spatial bands, and is therefore less susceptible to this aliasing
failure mode.

\begin{table}[h]
\centering
\caption{
Stochastic grid-shift ablation. For Allen--Cahn, removing the shifts produces
a classic collocation overfitting failure: the model attains a small residual
on the fixed training grid but fails between grid points. KdV is less affected
because the selected pyramid has lower spatial bandwidth.
}
\label{tab:shift_ablation}
\small
\begin{tabular}{llcc}
\toprule
Problem & Setting & Relative $L_2$ error & PDE residual loss \\
\midrule
Allen--Cahn & shifted & $1.52\times10^{-6}$ & $1.81\times10^{-9}$ \\
Allen--Cahn & no shift & $9.50\times10^{-1}$ & $2.12\times10^{-10}$ \\
KdV & shifted & $3.15\times10^{-5}$ & $1.12\times10^{-7}$ \\
KdV & no shift & $3.73\times10^{-5}$ & $9.65\times10^{-8}$ \\
\bottomrule
\end{tabular}
\end{table}

\subsection{Coarse-to-fine pyramid structure}
\label{sec:pyramid_fullres_ablation}

To isolate the effect of the pyramid structure from the effect of high
maximum resolution, we compare the Allen--Cahn pyramid used in the size sweep (see Figure~\ref{fig:pyramid_size_sweep})
with a parameter-matched ablation in which every level is fixed at the finest
spatial resolution, $512$.

\begin{table}[h]
\centering
\caption{
Allen--Cahn pyramid versus a parameter-matched all-full-resolution ablation.
Both models have comparable parameter counts and use the same finest spatial
resolution, but distributing capacity across resolutions gives substantially
lower error and residual than placing all feature grids at resolution $512$.
}
\label{tab:ac_pyramid_fullres}
\small
\begin{tabular}{@{}lccc@{}}
\toprule
Representation & Params & Relative $L_2$ error & PDE residual loss \\
\midrule
Pyramid, $14$ features/level & 411,938 & $1.52\times10^{-6}$ & $1.81\times10^{-9}$ \\
All levels at $512$, $5$ features/level & 419,586 & $1.32\times10^{-4}$ & $2.60\times10^{-8}$ \\
\bottomrule
\end{tabular}
\end{table}

The all-full-resolution ablation is $86.8\times$ worse in relative $L_2$ error
and has $14.4\times$ higher residual loss, despite a similar parameter count.
This shows that the improvement is not simply due to exposing high-resolution
Fourier features. The coarse-to-fine structure is a central part of the
embedding: it allocates capacity across spatial bands and optimizes
substantially better than stacking parameter-matched feature grids at the finest
resolution alone.

\subsection{Temporal pyramid blending}
\label{app:temporal_blending_ablation}

We ablate the temporal blending described in Section~\ref{sec:temporal_lerp}
using the Allen--Cahn \texttt{beignet} model. In the ablated model, each
pyramid level is collapsed to a single learned temporal slice. The raw time
coordinate is still passed to the decoder, so the ablation removes only the
learned temporal anchors and their linear blending.

\begin{table}[h]
\centering
\caption{
Allen--Cahn ablation with and without temporal pyramid blending. Both runs use
the same spatial pyramid, decoder, and optimizer. Temporal blending improves
both the final relative $L_2$ error and the PDE residual.
}
\label{tab:temporal_blending_ablation}
\small
\begin{tabular}{@{}llcc@{}}
\toprule
Setting & Temporal slices & Relative $L_2$ error & PDE residual loss \\
\midrule
Without temporal blending & 1 & $4.28\times10^{-6}$ & $3.38\times10^{-9}$ \\
With temporal blending & 8 & $1.52\times10^{-6}$ & $1.81\times10^{-9}$ \\
\bottomrule
\end{tabular}
\end{table}

Removing temporal blending increases the relative $L_2$ error by $2.81\times$
and the PDE residual by $1.86\times$. Thus, even though the decoder still
receives the raw time coordinate, the learned temporal anchors provide a useful
time-dependent feature representation for this Allen--Cahn configuration.

\section{Timing and efficiency comparison on Allen--Cahn}
\label{app:ac_timing_efficiency}

We compare the computational cost of Allen--Cahn residual training for three
settings: the FFT-structured \vibenet{} residual path, a pointwise DFT/JVP
evaluation of the same \vibenet{} model, and the PirateNet baseline
\citep{wang2024piratenets}. The goal of this experiment is to isolate
per-step residual-evaluation throughput from convergence behavior. We therefore
time raw optimizer updates after compilation at matched residual NFEs. Timings
exclude logging, evaluator calls, checkpointing, sampler generation during the
timed region, and periodic weight-recalculation steps. All timing tests were
performed on a single NVIDIA A100 80GB GPU.

\begin{table}[h]
\centering
\caption{
Raw Allen--Cahn training-step timing at matched residual NFEs. Values are
seconds per 100 optimizer steps, reported as mean $\pm$ standard deviation over
five timed trials. The FFT-structured \vibenet{} residual path is fastest at
every tested NFE.
}
\label{tab:ac_raw_step_timing}
\small
\begin{tabular}{@{}rccc@{}}
\toprule
Residual NFE & \textbf{\vibenet{} (FFT)} & \vibenet{} (DFT) & PirateNet \citep{wang2024piratenets} \\
\midrule
8,192  & \textbf{\boldmath$1.877 \pm 0.006$} & $7.601 \pm 0.012$ & $12.749 \pm 0.023$ \\
16,384 & \textbf{\boldmath$2.996 \pm 0.016$} & $14.582 \pm 0.122$ & $18.768 \pm 0.051$ \\
32,768 & \textbf{\boldmath$4.436 \pm 0.044$} & $30.097 \pm 0.096$ & $31.388 \pm 0.209$ \\
65,536 & \textbf{\boldmath$6.797 \pm 0.128$} & $55.561 \pm 0.427$ & $50.390 \pm 0.187$ \\
\bottomrule
\end{tabular}
\end{table}

The FFT path is substantially faster because spatial feature derivatives are
computed spectrally on structured grids and then reused across collocation
points. In contrast, the DFT/JVP variant and PirateNet baseline evaluate the
residual more pointwise, so their cost grows more directly with the number of
residual evaluations. At 65,536 residual NFEs, the FFT-structured path is
$8.2\times$ faster than the DFT/JVP evaluation of the same \vibenet{} model and
$7.4\times$ faster than PirateNet in raw training-step time.

Table~\ref{tab:ac_nfe_l2} reports the corresponding accuracy comparison. We
include the lowest-NFE PirateNet baseline and \vibenet{} runs obtained by
reducing the structured residual grid size. This separates the efficiency claim
from the accuracy claim: even at the matched 8,192 residual NFE setting,
\vibenet{} achieves lower final relative $L_2$ error than PirateNet, while the
larger structured-grid runs maintain similar accuracy at higher NFEs.

\begin{table}[h]
\centering
\caption{
Allen--Cahn final relative $L_2$ error for the lowest-NFE PirateNet baseline
and FFT-structured \vibenet{} runs at different residual NFEs. The matched
8,192-NFE comparison shows that the accuracy gain is not only due to using more
residual points.
}
\label{tab:ac_nfe_l2}
\small
\begin{tabular}{@{}lccc@{}}
\toprule
Method & $M_x$ & Residual NFE & Relative $L_2$ error \\
\midrule
PirateNet \citep{wang2024piratenets} & --- & 8,192 & $1.38\times10^{-5}$ \\
\textbf{\vibenet{} (FFT)} & 32  & 8,192  & \textbf{\boldmath$1.91\times10^{-6}$} \\
\textbf{\vibenet{} (FFT)} & 64  & 16,384 & \textbf{\boldmath$1.32\times10^{-6}$} \\
\textbf{\vibenet{} (FFT)} & 128 & 32,768 & \textbf{\boldmath$1.78\times10^{-6}$} \\
\textbf{\vibenet{} (FFT)} & 256 & 65,536 & \textbf{\boldmath$1.52\times10^{-6}$} \\
\bottomrule
\end{tabular}
\end{table}

These results show two complementary benefits of the structured residual path.
First, at a fixed residual NFE, FFT-based derivative evaluation gives much
higher training-step throughput than pointwise residual evaluation. Second, the
memory efficiency of the structured path makes it practical to train with many
more collocation points: the $65{,}536$-NFE \vibenet{} run fits on a single
NVIDIA A100 80GB GPU, whereas the PirateNet baseline requires two GPUs for its
$8{,}192$-NFE configuration with gradient-norm loss reweighting. Thus the
structured \vibenet{} residual path improves both the matched-NFE comparison and
the feasible high-NFE training regime.

\section{Spectral Error Analysis}

\begin{figure}[t]
  \centering
  \includegraphics[width=\linewidth]{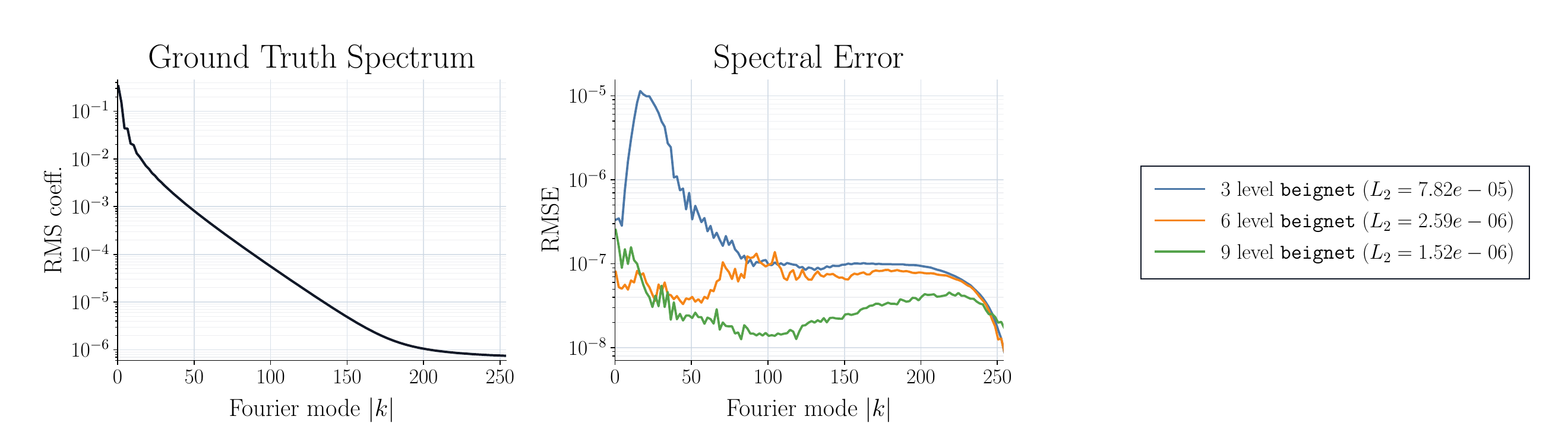}
  \caption{
  Frequency-resolved Allen--Cahn diagnostic. Left: root mean squared spatial
  Fourier coefficient magnitude of the ground-truth solution, averaged over
  time. Right: root mean squared spatial Fourier coefficient magnitude of the
  prediction error for 3-, 6-, and 9-level \vibenet{} models. The duplicated
  periodic endpoint in the reference grid is removed before applying the FFT,
  and adjacent Fourier modes are RMS-binned in pairs.
  }
  \label{fig:spectral_error}
\end{figure}

Fig. \ref{fig:spectral_error} shows a frequency-resolved diagnostic for the
Allen--Cahn size sweep.  The left panel reports the spectrum of the reference
solution, and the right panel reports the spectrum of the prediction error for
3-, 6-, and 9-level \vibenet{} models. For each selected checkpoint, we evaluate
the learned solution on the Allen--Cahn reference grid and compute the spatial
Fourier transform of the error field
\[
  e(t,x) = u_\vct{\theta}(t,x) - u_{\rm ref}(t,x).
\]
The plotted value at mode $|k|$ is the root mean squared Fourier coefficient
magnitude over time, binned over neighboring pairs of Fourier
modes. The comparison isolates how increasing the Fourier feature pyramid bandwidth
changes the error distribution across spatial modes.  The 3-level model has
limited spatial bandwidth and leaves substantially larger error in the
low modes.  Increasing to 6 levels reduces the error by roughly an order
of magnitude across most of the spectrum, while the 9-level model further lowers
the error, especially in the higher-frequency tail.  This supports the
observation in Fig. \ref{fig:pyramid_size_sweep} that Allen--Cahn benefits from a
wider pyramid: its sharp interface dynamics require resolving a broad range of
spatial harmonics, and increasing the maximum represented feature resolution
reduces the spectral error rather than only improving low-frequency fit.

\begin{table}[p]
  \centering
  \caption{
  Configuration comparison for the \vibenet{} models used in the JAX-PI
  benchmark experiments. 
  }
  \label{tab:vibenet_configurations}
  \scriptsize
  \setlength{\tabcolsep}{3pt}
  \renewcommand{\arraystretch}{1.08}
  \resizebox{\linewidth}{!}{%
  \begin{tabular}{llcccc}
    \toprule
    Area & Field & Allen--Cahn & KdV & Ginzburg--Landau & Gray--Scott \\
    \midrule
    Pyramid & \texttt{num\_scales} & 9 & 4 & 5 & 5 \\
    Pyramid & \texttt{num\_features} per level & 14 & 64 & 64 & 64 \\
    Pyramid & Min grid & \texttt{x=2, t=8} & \texttt{x=2, t=8} & \texttt{x=2, y=2, t=8} & \texttt{x=2, y=2, t=8} \\
    MLP & Width x depth & \texttt{256 x 4} & \texttt{256 x 4} & \texttt{128 x 3} & \texttt{128 x 3} \\
    MLP & Activation & \texttt{tanh} & \texttt{tanh} & \texttt{swish} & \texttt{swish} \\
    MLP & Reparam & \texttt{weight\_fact}, mean 1.0, std 0.1 & \texttt{weight\_fact}, mean 1.0, std 0.1 & \texttt{weight\_fact}, mean 0.5, std 0.1 & \texttt{weight\_fact}, mean 0.5, std 0.1 \\
    Training & \texttt{max\_steps} & 300,000 & 200,000 & 100,000 per window & 100,000 per window \\
    Training & Time windows & not configured & not configured & 5 & 10 \\
    Optim & Optimizer & Adam & Adam & Adam & Adam \\
    Optim & LR & \texttt{1e-3} & \texttt{1e-3} & \texttt{1e-3} & \texttt{1e-3} \\
    Optim & Decay steps & 3000 & 2000 & 2000 & 2000 \\
    Optim & Warmup steps & 1200 & 0 & 5000 & 0 \\
    \bottomrule
  \end{tabular}%
  }
\end{table}

\section{KdV is stiff at high frequencies}
\label{app:kdv_stiff_high_frequencies}

The KdV size sweep in Fig.~\ref{fig:pyramid_size_sweep} favors a compact
pyramid rather than the widest representation. This is consistent with the
structure of the KdV residual, whose spatial part contains a third derivative:
high-frequency Fourier modes are strongly amplified by the PDE operator. Thus,
for KdV, adding finer pyramid levels does not merely increase representational
bandwidth; it can also introduce trainable directions that are poorly
conditioned for residual minimization.

\begin{figure}[t]
  \centering
  \includegraphics[width=\linewidth]{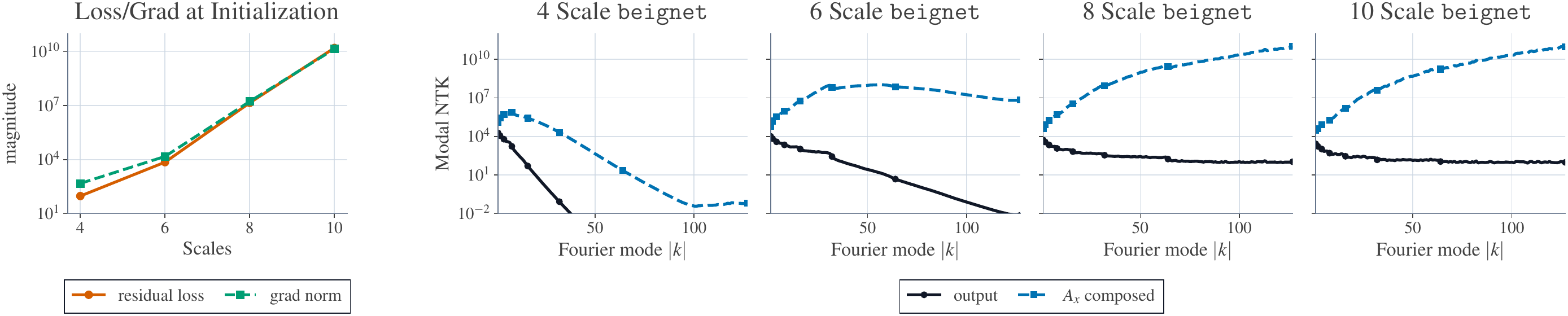}
  \caption{
  Initialization diagnostic for KdV across 4-, 6-, 8-, and 10-scale
  \texttt{beignet} models. The loss/gradient panel reports initial residual loss
  and pyramid-gradient norm, both of which grow rapidly as finer scales are
  added. The modal panels compare the model-output modal NTK with the same
  tangent response composed with the linearized KdV spatial residual operator $A_x$. Increasing the number of scales exposes higher represented frequencies whose operator-composed
  response grows rapidly, matching the rise in initialization loss and gradient
  magnitude; this results in optimization instability at higher frequency levels of the pyramid.
  }
  \label{fig:kdv_stiff_init_ntk}
\end{figure}

Figure~\ref{fig:kdv_stiff_init_ntk} first compares the initial residual loss and
pyramid-gradient norm across KdV pyramids with increasing numbers of scales.
Both quantities grow rapidly as finer levels are added, suggesting that the
wider pyramids are already poorly conditioned at initialization.

To explain this effect, we additionally propose a rough modal tangent diagnostic. For a spatial Fourier
probe $q_k(x)$, such as a sine or cosine mode at wavenumber $k$, define
$a_k(\vct{\theta})=\langle u_{\vct{\theta}},q_k\rangle$. The squared norm
$\|\nabla_{\vct{\theta}} a_k\|_2^2$ gives the tangent energy associated with
that Fourier mode. This is a finite-width empirical analogue of the NTK
spectral view used to study frequency-dependent learning in coordinate
networks: when the kernel is approximately stationary, Fourier modes
diagonalize it, and the corresponding NTK eigenvalues control the convergence
rates of individual frequency components~\citep{tancik2020fourier,cao2019towards}.

We also compute an operator-composed variant by applying the KdV spatial
residual operator, linearized around the initial condition $u_0$,
\[
  A_x[v] = u_0\, v_x + (\partial_x u_0)\, v + 0.022^2\, v_{xxx},
\]
before taking the parameter tangent, replacing
$a_k(\vct{\theta})$ with
$\langle A_x[u_{\vct{\theta}}], q_k\rangle$. This curve measures
the frequency-dependent sensitivity seen by the residual objective, rather than
the sensitivity of the raw model output alone.

The modal tangent curves show that the rise in initialization loss and gradient
norm is not explained by the raw output sensitivity alone. The model-output
modal NTK remains comparatively moderate, while the operator-composed response
grows by many orders of magnitude as finer pyramid levels expose higher
frequencies. This supports the interpretation that broad KdV pyramids become
optimization-stiff because high-frequency degrees of freedom are strongly
amplified by the residual operator. The best compact model instead allocates
capacity to low- and mid-frequency structure without exposing as much stiff
high-frequency response.

\section{\vibenet{} configurations for each problem}
\label{app:vibenet_configurations}

Table~\ref{tab:vibenet_configurations} summarizes the configuration settings we used for the four JAX-PI benchmark PDEs.

\section{JAX-PI Benchmark Descriptions}
\label{app:jaxpi_benchmarks}

The four time-dependent benchmarks follow the JAX-PI\footnote{\url{https://github.com/PredictiveIntelligenceLab/jaxpi}}
\citep{wang2023expert} repository. Training minimizes an initial-condition loss and a PDE
residual loss on uniformly sampled collocation points, with causal time-chunk
weighting and periodic loss-weight updates. Evaluation reports relative $L_2$
error on the stored solver grid. Our \vibenet{} runs use the same entry points
and data files, replacing the coordinate embedding and residual-evaluation
path (and, for the 2D systems, the MLP width); the two 2D systems are trained in time windows, passing the predicted
terminal state from one window as the next initial condition.

\paragraph{Allen--Cahn.}
The Allen--Cahn equation is
\begin{equation}
  u_t - 10^{-4} u_{xx} + 5u^3 - 5u = 0,
  \qquad x \in [-1,1],\quad t\in[0,1],
\end{equation}
with periodic boundary conditions and initial condition
$u(x,0)=x^2\cos(\pi x)$. This benchmark contains sharp phase-interface
dynamics and benefits from broad spatial spectral coverage. The reference file is generated with MATLAB Chebfun's \texttt{spin} solver on
201 time samples and a 512-point periodic spatial grid, including the repeated
endpoint.

\paragraph{KdV.}
The Korteweg--de Vries equation is
\begin{equation}
  u_t + u u_x + 0.022^2 u_{xxx}=0,
  \qquad x \in [-1,1],\quad t\in[0,1],
\end{equation}
with periodic boundary conditions and initial condition $u(x,0)=\cos(\pi x)$.
Unlike Allen--Cahn, KdV contains a third-order spatial derivative, so high
spatial modes are strongly amplified in the residual. The reference file is generated with MATLAB Chebfun's \texttt{spin} solver on 201 time
samples and the same 512-point periodic spatial grid.

\paragraph{Complex Ginzburg--Landau.}
The complex Ginzburg--Landau equation is a two-component reaction--diffusion
system for $A=u+iv$ on the periodic domain $\mathbf{x}=(x,y)\in[-1,1]^2$. In real
variables, the system is
\begin{align}
  u_t &= \varepsilon (u_{xx}+u_{yy})
  + \kappa\left[u-u(u^2+v^2)+\frac{3}{2}v(u^2+v^2)\right], \\
  v_t &= \varepsilon (v_{xx}+v_{yy})
  + \kappa\left[v-v(u^2+v^2)-\frac{3}{2}u(u^2+v^2)\right].
\end{align}
The cubic reaction terms and cross-coupling generate coherent wave and defect
dynamics, providing a 2D test of the multi-resolution representation. The
reference uses $\kappa=10$, $\varepsilon=\kappa/50^2$, 101 time samples, and a
$200\times200$ spatial grid; training is split into five time windows.

\paragraph{Gray--Scott.}
The Gray--Scott system is a two-component autocatalytic reaction--diffusion
model on the periodic domain $\mathbf{x}=(x,y)\in[-1,1]^2$:
\begin{align}
  u_t &= \varepsilon_u (u_{xx}+u_{yy}) + b_1(1-u) - c_1uv^2, \\
  v_t &= \varepsilon_v (v_{xx}+v_{yy}) - b_2v + c_2uv^2 .
\end{align}
It produces fine-scale spot and labyrinthine patterns, making it a stringent
test of spatial resolution and multi-scale feature capacity. The reference uses
$(\varepsilon_u,\varepsilon_v,b_1,b_2,c_1,c_2)=(0.2,0.1,40,100,1000,1000)$,
101 time samples over $t\in[0,2]$, and a $200\times200$ spatial grid; training
is split into ten time windows.

\section{Self-similar Burgers blowup details}
\label{app:burgers_blowup_details}

We follow the 1D Burgers self-similar profile formulation of
\citet{wang2026high}.  For fixed similarity exponent $\lambda=0.5$, the profile
$U$ satisfies
\begin{equation}
    F_U(y)
    :=
    -\lambda U(y)
    +
    \bigl((1+\lambda)y+U(y)\bigr)U_y(y)
    =
    0 .
    \label{eq:app_burgers_residual}
\end{equation}
Since the profile is defined on an unbounded domain, we train in the
computational coordinate $\eta\in[0,c]$ with $c=30$ and map to the profile
coordinate by $y=\sinh(\eta)$.  Derivatives in Eq.~\eqref{eq:app_burgers_residual}
are therefore evaluated with the chain-rule factor
$\partial_y=(\cosh\eta)^{-1}\partial_\eta$.  We use the same analytic structure
as in \citet{wang2026high} to build the known symmetry and far-field behavior
into the ansatz:
\begin{equation}
    U(y(\eta))
    =
    \widetilde U_\vct{\theta}(\eta)-\widetilde U_\vct{\theta}(-\eta)
    -
    \left(\frac{y}{1+y}\right)^{15}
    y^{\lambda/(1+\lambda)},
    \qquad
    y=\sinh(\eta),
    \label{eq:app_burgers_repr}
\end{equation}
where $\widetilde U_\vct{\theta}$ is the learnable component.  This construction
enforces odd symmetry through antisymmetrization and incorporates the prescribed
leading-order tail through the final analytic term.

The Adam runs sample a fresh batch $\eta\sim\mathrm{Unif}(0,c)$ at each step and
minimize the squared profile residual together with the squared derivative
residual,
\begin{equation}
    \mathcal L
    =
    \mathbb E_{\eta\sim \mathrm{Unif}(0,c)}
    \left[
        |F_U(y(\eta))|^2
        +
        |\partial_\eta F_U(y(\eta))|^2
    \right].
    \label{eq:app_burgers_loss}
\end{equation}
Both Adam rows use JAX with 64-bit arithmetic, 20K steps, batch size 4096, Adam
with learning rate $10^{-4}$, cosine decay to zero, $\epsilon=10^{-15}$, no
weight decay, and seed 0.  The coordinate MLP has width 20, depth 4, and tanh
activations.  The \vibenet{} model uses the same decoder MLP and maps $\eta$ to
$(\eta+c)/(2c)$ before querying a 12-level Fourier feature pyramid with 4
channels per level, grid sizes $2,4,\ldots,4096$, global preconditioner 1000,
per-level preconditioner 0.30, and initialization noise $10^{-5}$.

The BFGS row uses the same width-20, depth-4 coordinate MLP, trained in PyTorch
float64 on CUDA with batch size 10000, seed 0, and 20000 BFGS iterations in
blocks of 1000 using the SSBroyden1 inverse-Hessian update.  Its training
objective follows the high-precision reference implementation and includes
$0.1\,\mathbb E|F_U|^2+\mathcal L_b+0.1\,\mathbb E|\partial_\eta F_U|^2$ with a
boundary term $\mathcal L_b$.  For Table~\ref{tab:burgers}, however, the saved
BFGS checkpoint is re-evaluated with the same unweighted diagnostic definitions
used for the JAX rows.

All table metrics are evaluated after training on the centered diagnostic grid
$\eta_i=c(i+1/2)/1000$, $i=0,\ldots,999$, with $y_i=\sinh(\eta_i)$.  We report
$1000^{-1}\sum_i F_U(y_i)^2$ and $\log_{10}\max_i |F_U(y_i)|$; this grid is used
only for reporting and is not resampled during optimization.


\end{document}